\documentclass[sigconf]{acmart}
\usepackage{enumitem}
\usepackage{multirow}
\usepackage{booktabs}
\usepackage{array}
\usepackage{graphicx}
\usepackage{adjustbox}
\usepackage{arydshln}

\AtBeginDocument{%
  }

\copyrightyear{2025}
\acmYear{2025}
\setcopyright{cc}
\setcctype{by}
\acmConference[WWW Companion '25] {Companion Proceedings of the ACM Web Conference 2025}{April 28-May 2, 2025}{Sydney, NSW, Australia.}
\acmBooktitle{Companion Proceedings of the ACM Web Conference 2025 (WWW Companion '25), April 28-May 2, 2025, Sydney, NSW, Australia}
\acmISBN{979-8-4007-1331-6/25/04}
\acmDOI{10.1145/3701716.3715539}

\begin{document}

\title{Panoramic Interests: Stylistic-Content Aware Personalized~Headline~Generation}

\author{Junhong~Lian}
\authornotemark[2]
\authornotemark[5]
\orcid{0000-0002-2922-1216}
\affiliation{%
  \institution{Institute of Computing Technology, Chinese Academy of Sciences}
  \state{Beijing}
  \country{China}
} 
\email{lianjunhong23s@ict.ac.cn}

\author{Xiang~Ao}
\authornote{Corresponding authors.}
\authornote{Key Lab of Intelligent Information Processing, Institute of Computing Technology (ICT), Chinese Academy of Sciences (CAS). Also affiliated with the Key Lab of AI Safety, CAS, Beijing, China.}
\authornote{Institute of Intelligent Computing Technology, CAS, Suzhou, China.}
\authornotemark[5]
\orcid{0000-0001-9633-8361}
\affiliation{%
  \institution{Institute of Computing Technology, Chinese Academy of Sciences}
  \state{Beijing}
  \country{China}
}
\email{aoxiang@ict.ac.cn}

\author{Xinyu~Liu}
\authornotemark[1]
\authornote{High Performance Computer Research Center, ICT, CAS.}
\authornote{The authors are also with the University of Chinese Academy of Sciences, CAS.}
\orcid{0009-0001-9481-5758}
\affiliation{%
  \institution{Institute of Computing Technology, Chinese Academy of Sciences}
  \state{Beijing}
  \country{China}
}
\email{liuxinyu@ict.ac.cn}

\author{Yang~Liu}
\authornotemark[2]
\authornotemark[5]
\orcid{0000-0002-1525-0788}
\affiliation{%
  \institution{Institute of Computing Technology, Chinese Academy of Sciences}
  \state{Beijing}
  \country{China}
} 
\email{liuyang2023@ict.ac.cn}

\author{Qing~He}
\authornotemark[2]
\authornotemark[5]
\orcid{0000-0001-8833-5398}
\affiliation{%
  \institution{Institute of Computing Technology, Chinese Academy of Sciences}
  \state{Beijing}
  \country{China}
}
\email{heqing@ict.ac.cn}

\renewcommand{\shortauthors}{Junhong Lian, Xiang Ao, Xinyu Liu, Yang Liu, and Qing He}

\begin{abstract}
  Personalized news headline generation aims to provide users with attention-grabbing headlines that are tailored to their preferences.
  Prevailing methods focus on user-oriented content preferences, but most of them overlook the fact that diverse stylistic preferences are integral to users' panoramic interests, leading to suboptimal personalization. 
  In view of this, we propose a novel \textbf{\underline{S}}tylistic-\textbf{\underline{C}}ontent \textbf{\underline{A}}ware \textbf{\underline{Pe}}rsonalized Headline Generation (SCAPE) framework. 
  SCAPE extracts both content and stylistic features from headlines with the aid of large language model~(LLM) collaboration.
  It further adaptively integrates users' long- and short-term interests through a contrastive learning-based hierarchical fusion network.
  By incorporating the panoramic interests into the headline generator, SCAPE reflects users' stylistic-content preferences during the generation process.
  Extensive experiments on the real-world dataset PENS demonstrate the superiority of SCAPE over baselines.
\end{abstract}



\maketitle

\begin{figure}[!t]
  \centering
  \includegraphics[width=0.95\linewidth]{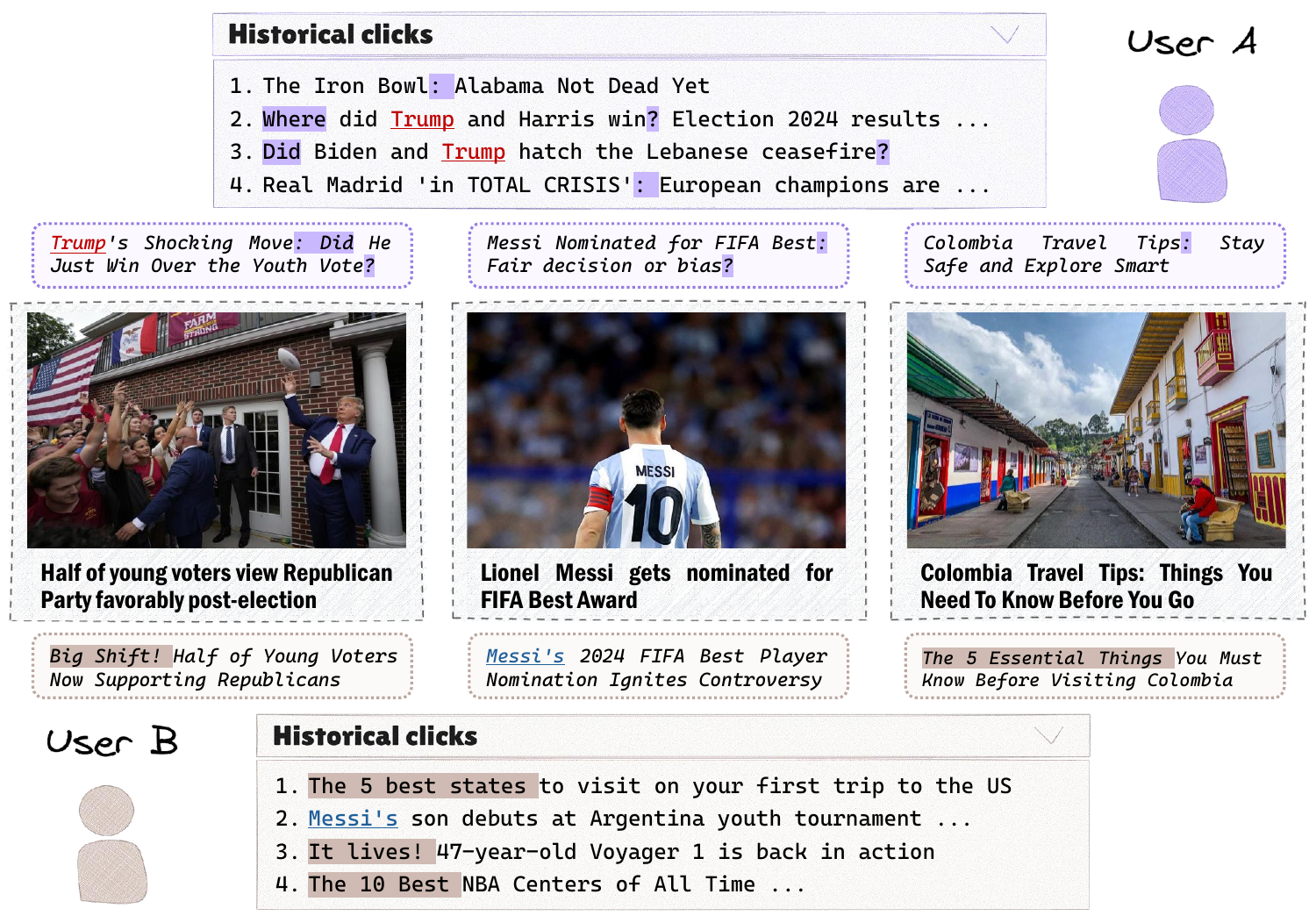}
  \caption{Illustration of the Joint Influence of Content Interests and Stylistic Preferences on Headline Personalization.}
  \Description{The illustration demonstrates how content interests and stylistic preferences jointly impact headline personalization for two users, A and B. Both users' personalized headlines reflect their unique content interests and stylistic preferences.}
  \label{fig:intro}
  \vspace{-1.5em}
\end{figure}
\section{Introduction}
Generating attractive news headlines has been firmly established as a special form of text summarization~\cite{luo2019reading, gu2020generating}.
Numerous efforts have focused on generating attention-grabbing headlines through personalized approaches, recognizing that readers with diverse preferences may find different focal points in the same news~\cite{ao2021pens, ao2023put}.
These methods aimed to build engaging headlines tailored to individual users' reading interests by utilizing auxiliary information, such as user profiles and historical clicks~\cite{zhang2022personalized, yang2023fact, song2023general, tan2024enhancing}.
In spite of that, existing personalized approaches primarily emphasize user-oriented content-driven headlines, while largely overlooking the stylistic features of news headlines.

The stylistic features of news headlines are pivotal in journalism communication~\cite{bell1991language, fiske2010introduction}. Boosting headline engagement through text style transfer has become a widely-used approach~\cite{jin2020hooks, zhang2023mediahg}. Style transfer-based methods incorporate specific stylistic elements or employ interrogative forms in news headlines to capture readers' attention~\cite{shu2018deep, jin2020hooks}. 
Nevertheless, these style transfer-based methods rely on a single, uniform stylistic strategy to boost engagement risks crossing into clickbait~\cite{ao2023put, yang2023fact}. To improve content consistency, recent works~\cite{li2021style, zhang2023mediahg} have emphasized the style-content duality in headline generation, focusing on transferring to prototype headline styles while maintaining content decoupling. 
Despite these efforts, style transfer-based methods are unsatisfactory because they have yet to achieve user-oriented personalization.


In this work, we focus on encompassing both content interests and stylistic preferences to fully meet users' needs for personalized headline generation.
The reason is that even readers with similar content preferences may exhibit distinct style preferences for headlines due to different personal reading habits.
For example, consider the case illustrated in Figure~\ref{fig:intro}, wherein users' content focus and stylistic preferences jointly affect the personalization of headlines.
We mark their content interests and stylistic patterns with different colors and text highlighting.
Base on \textsf{User A}'s historical click patterns, we can infer that \textsf{User A} prefers the \textsf{\textit{``subject-subordinate title''}} format and interrogative forms to increase engagement, while \textsf{User B} favors \textsf{\textit{``numbered list''}} headlines and short exclamatory sentences.
Hence, we contend that both content interests and stylistic preferences are interwoven to form users' panoramic interests in personalization and must be considered simultaneously. 
However, simultaneously addressing this issue is challenging. A primary obstacle is how to extract the inherent content and stylistic features encapsulated in headlines without ground-truth labels.
Moreover, another challenge is to decouple implicit user panoramic interests and integrate them into headlines without explicit supervision.

To remedy the above challenges, we propose a novel framework named \textbf{\underline{S}}tylistic-\textbf{\underline{C}}ontent \textbf{\underline{A}}ware \textbf{\underline{Pe}}rsonalized Headline Generation~(SCAPE).
We first designed a headline inference module that uses LLM to infer content and stylistic attributes for each news headline individually.
It subsequently produces an offline embedding table that encapsulates these attributes at the headline-level.
Furthermore, a hierarchical gated fusion network is devised to adaptively integrate users' long- and short-term content interests and stylistic preferences. 
In addition, a self-supervised strategy is adopted to decouple the intertwined user preferences.
Lastly, a personalized injection module leverages the fused user representation to guide a lightweight generator to produce final outputs. This ensures that the generated headlines reflect user's panoramic interests.  
Our main contributions are summarized as follows:
\begin{itemize}[leftmargin=2em]
    \item To the best of our knowledge, this work represents the first attempt to incorporate both content and style preferences from user profiles for personalized news headline generation.
    \item We introduce a novel framework, SCAPE, which extracts the inherent features of headlines through LLM collaboration and facilitates personalized generation via a hierarchical fusion of both content interests and stylistic preferences.
    \item Extensive evaluations conducted on the real-world personalized news headline generation benchmark PENS~\cite{ao2021pens} demonstrate the superior performance of our method.
\end{itemize}

\section{Methodology}

\subsection{Problem Formulation}
Consider a news database denoted as \( \mathcal{D} = \{n_i = (t_i, b_i)\}_{i=1}^{|\mathcal{D}|} \), where \( t_i \) and \( b_i \) represent the original headline and the body of news \( n_i \), and \( |\mathcal{D}| \) is the total number of news.
For a given user \( u \), we denoted \( u \)'s click history as \( H_u = [t_{h_1}, t_{h_2}, \cdots, t_{h_L}] \), comprising \( L \) clicked headlines \( t_{h_j} \) (\( j = 1, \cdots, L \)), where each \( t_{h_j} \) satisfies \( t_{h_j} \in \{t_i \mid n_i = (t_i, b_i) \in \mathcal{D}\} \).
Then, given a candidate news \( n_c = (t_c, b_c) \in \mathcal{D} \), our target is to generate a personalized headline \( P_{c}^{u} \) for user \( u \) based on \( u \)'s click history \( H_u \) and the body \( b_c \) of the candidate news.

\begin{figure}[!t]
  \centering
  \includegraphics[width=0.95\linewidth]{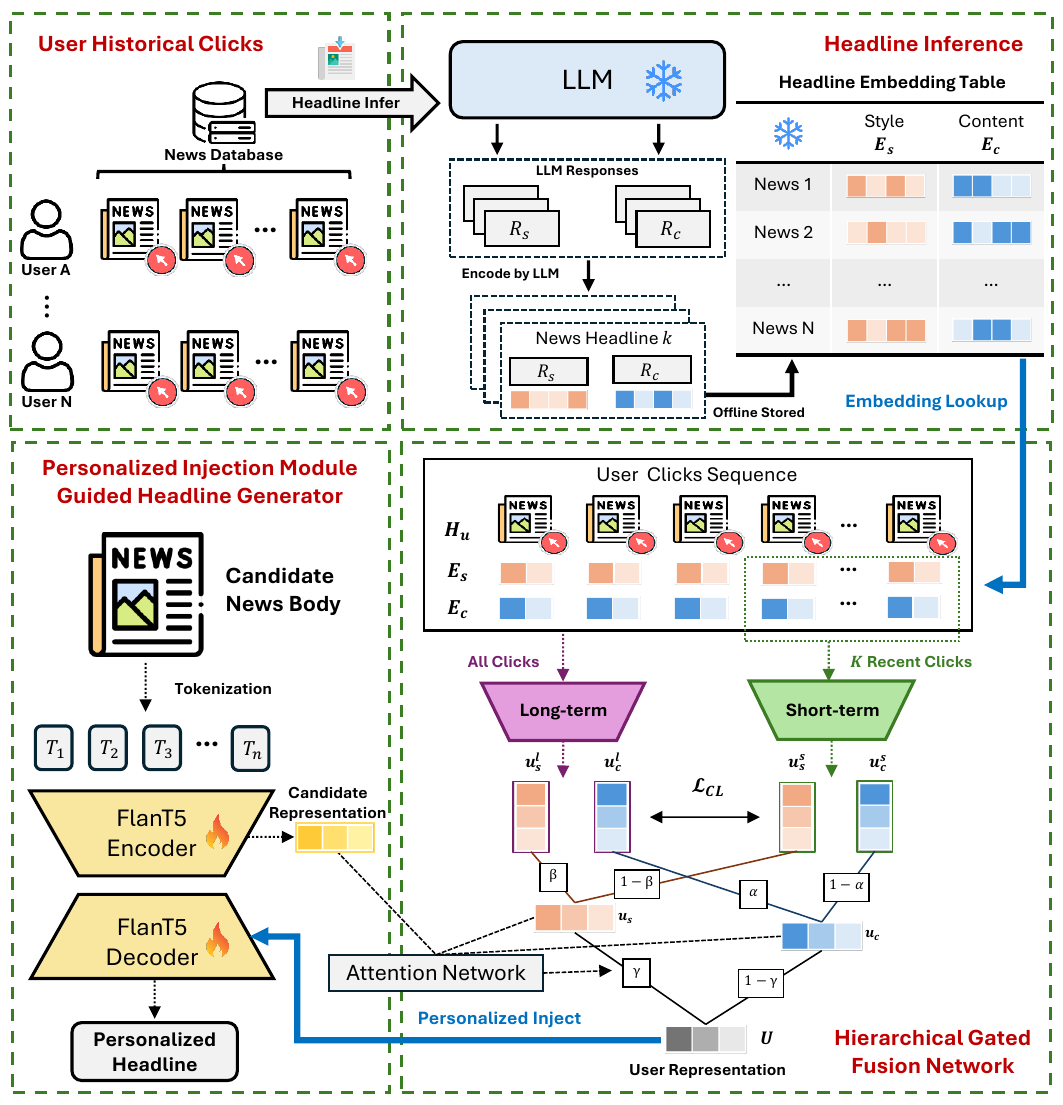}
  \caption{Architecture of the Proposed SCAPE Framework.}
  \Description{Architecture of the Proposed SCAPE Framework.}
  \label{fig:model}
  \vspace{-1em}
\end{figure}
\subsection{Our SCAPE Framework}
In this section, we describe our proposed SCAPE framework, which is illustrated in Figure~\ref{fig:model}. It consists of a headline inference module, a hierarchical gated fusion network, and a personalized injection module for headline generator.

\subsubsection{Headline Inference}

Extracting the inherent features from headlines without ground-truth labels is challenging, but LLMs with extensive world knowledge offer a promising solution for inferring high-level latent concepts.
Building on this, the headline inference module in SCAPE employs an instruction-tuned LLM, denoted as $LLM_{inst}$, to infer text style and content interests from headlines $t_i$ in news database $\mathcal{D}$ as follows:
\begin{equation}
  \small
  R_s = LLM_{inst}(t_i, \mathcal{P}_{\text{style}}) \quad 
  R_c = LLM_{inst}(t_i, \mathcal{P}_{\text{content}}) 
  \label{eq:response}
\end{equation}
where $\mathcal{P}_{\text{style}}$ and $\mathcal{P}_{\text{content}}$ are the prompts designed to instruct the LLM to infer stylistic features and content interests, respectively, with $R_s$ and $R_c$ as corresponding responses.

Inspired by instruction-following text embedding~\cite{peng2024answer}, we design task-specific instructions $I_s$ and $I_c$ for embedding style and interest. 
These instructions are concatenated with news headlines and their corresponding responses, which are then encoded by an embedding-based LLM, denoted as $LLM_{emb}$.
The offline stored headline embedding table $E$ is constructed as follows:
\begin{equation}
  \small
  E_s = LLM_{emb}([t_i, I_s, R_s]) \quad 
  E_c = LLM_{emb}([t_i, I_c, R_c]) 
  \label{eq:embedding}
\end{equation}
\begin{equation}
  \small
  E = \{(E_s(t_i), E_c(t_i)) \mid n_i = (t_i, b_i) \in \mathcal{D}\}
  \label{eq:embed-table}
\end{equation}
where $E_s(t_i)$ and $E_c(t_i)$ represent the style and content embeddings of headline $t_i$, respectively, and $E$ is the set of style-content embedding pairs for all headlines in $\mathcal{D}$.

\subsubsection{Hierarchical Stylistic-Content Awareness Fusion}
\label{sec:Fusion}

Given that historical clicks reflect both stable personal traits and recent short-term characteristics, we employ an attention mechanism to aggregate all clicks and a GRU network for the $K$ recent clicks. 
This yields the user's long-term content interests representation $u_{c}^{l}$ and short-term content interests representation $u_{c}^{s}$ as follows: 
\begin{equation}
  \small
  \tilde{E}_c(t_{h_j}) = \text{MLP}(E_c(t_{h_j})) \quad \forall j \in \{1, \dots, L\}
\end{equation}
\begin{equation}
  \small
  u_{c}^{l} = \text{Attn}(\tilde{E}_c(t_{h_1}), \dots, \tilde{E}_c(t_{h_L}))
\end{equation}
\begin{equation}
  \small
  u_{c}^{s} = \text{GRU}([ \tilde{E}_c(t_{h_{L-K+1}}), \dots, \tilde{E}_c(t_{h_L}) ])
\end{equation}
where $\tilde{E}_c(t_{h_j})$ is the projected content embedding for the $j$-th historical click with $L > K$. Similarly, we obtain the user's long-term and short-term stylistic preferences, denoted as $u_{s}^{l}$ and $u_{s}^{s}$.

We then obtain the fusion of long- and short-term representations corresponding to different types of interests, as follows:
\begin{equation}
  \small
  \alpha = \sigma(W_{gc} \cdot [u_{c}^{l}, u_{c}^{s}] + b_{gc})
  \qquad
  \beta = \sigma(W_{gs} \cdot [u_{s}^{l}, u_{s}^{s}] + b_{gs})
  \label{eq:content_fusion_alpha}
\end{equation}
\begin{equation}
  \small
  u_{c} = \alpha \cdot u_{c}^{l} + (1 - \alpha) \cdot u_{c}^{s}
  \qquad
  u_{s} = \beta \cdot u_{s}^{l} + (1 - \beta) \cdot u_{s}^{s}
  \label{eq:content_fusion_uc}
\end{equation}
where $\sigma$ denotes the sigmoid activation function, and $\alpha$ is the gating weight for content interests. $W_{gc}$ and $b_{gc}$ are learnable parameters. 
Similarly, we derive the gating weight $\beta$ and representation $u_{s}$ for style preferences through the same process.

Subsequently, we integrate the $u_{c}$ and $u_{s}$ based on the attention weights derived from the candidate article representation. This stylistic-content awareness fusion mechanism enables the stepwise integration of diverse representations of the user. 

\subsubsection{Personalized Injection Module Guided Generator}
Once the user representation $U$ is obtained, SCAPE integrates it into a decoder of the lightweight headline generator through the personalized injection module. Specifically, the user representation $U$ is added to the input embeddings $\mathbf{X}$ of each token in the decoder. 
This user-specific vector is subsequently propagated through the residual flow, thereby influencing the generated headlines. By integrating user preferences at the token level, the headline generator can better align its output with the user's interests, producing stylistic-content-aware personalized headlines that effectively reflect their panoramic interests, formulated as follows:
\begin{equation}
  \small
  \mathbf{X}' = \mathbf{X} + (\mathbf{1}_n \otimes U)
\label{eq:injection}
\end{equation}
where $\mathbf{X} \in \mathbb{R}^{n \times d}$, $U \in \mathbb{R}^d$, $n$ is the sequence length, $d$ is the embedding dimension, and $\mathbf{1}_n \in \mathbb{R}^n$ is an all-ones vector.


\subsection{Disentanglement Strategy}
To prevent the $u_{c}$ and $u_{s}$ in Section~\ref{sec:Fusion} from collapsing into trivial forms, we adopt a mechanism inspired by~\cite{zheng2022disentangling} to promote the disentanglement of long- and short-term (LS-term) style and content interest representations.
Specifically, we aggregate users' historical LS-term clicks using mean aggregation to derive proxies for stylistic preferences ($p_{u_{s}}^{l}$ and $p_{u_{s}}^{s}$) and content interests ($p_{u_{c}}^{l}$ and $p_{u_{c}}^{s}$).

We employ contrastive learning between user LS-term representations and proxies, ensuring that the learned representations of LS-term style or content are more similar to their respective proxies than to opposite proxies.
For content interests modeling, a user's long-term content interest $u_{c}^{l}$ should be more similar to the long-term content proxy $p_{u_{c}}^{l}$ than to the short-term content proxy $p_{u_{c}}^{s}$ or the long-term style proxy $p_{u_{s}}^{l}$.

The triplet loss function is used to achieve contrastive learning, and the final contrastive learning loss can computed as follows:
\begin{equation}
  \small
    \mathcal{L}_{\text{CL}} = \frac{1}{|\mathcal{R}|} \sum_{R \in \mathcal{R}} \frac{1}{|\mathcal{N}_R|} \sum_{N \in \mathcal{N}_R} \max \left( 0, \ d(R, P_R) - d(R, N) + m \right )
    \label{eq:cl_loss} 
\end{equation}
where $\mathcal{R} = \{ u_{s}^{l}, u_{s}^{s}, u_{c}^{l}, u_{c}^{s} \}$ is user representations set, and $\mathcal{N}_R$ is negative proxies set. $P_R$ is the positive proxy corresponding to $R$. $d$ denotes the Euclidean distance, and $m$ is a positive margin value.

\section{Experiments}
\subsection{Experimental Setup}

\subsubsection{Datasets and Baselines}
We use the publicly available personalized news headline generation dataset PENS~\cite{ao2021pens} as benchmark. The dataset comprises $500,000$ anonymized impressions from over $445,000$ users and $113,762$ news articles, capturing detailed historical click data to reflect nuanced personalized preferences. The test set includes $3,940$ news items annotated by $103$ users, each providing $200$ unique parallel headlines as the gold standard for personalized headline evaluation.

We compare SCAPE with SOTA personalized headline generation methods on the PENS benchmark, including pointer-network-based frameworks~\cite{ao2021pens, zhang2022personalized, ao2023put} and methods based on pre-trained language models~\cite{yang2023fact, song2023general}. 
We also expanded our evaluation by employing a prompt-based method to comprehensively investigate the performance of LLMs against strong personalized baselines, evaluating both open-source LLMs~\cite{chung2024scaling, qwen2} of various sizes and closed-source LLMs~\cite{glm2024chatglm, liu2024deepseek, hurst2024gpt} via API services.

\subsubsection{Evaluation Metrics}
We use ROUGE metrics to measure lexical similarity, with ROUGE-1 and ROUGE-2 for informativeness, and ROUGE-L for fluency.
To evaluate the factual consistency of generated headlines, we follow previous work~\cite{yang2023fact} that reported Fact Scores. 
As for personalization, due to the lack of a widely accepted evaluation method, we design a pairwise comparison task~\cite{wang2024large}, where strong LLMs evaluate candidates against original headlines based on the user's historical clicks. To minimize bias, we swap the contextual order of candidates and perform two independent assessments, marking inconsistencies as ties. Personalization performance is reported as "win/tie/lose" outcomes between the candidates and original headlines across baseline methods.

\subsubsection{Implementation Details}
We use FlanT5-base as the backbone for the headline generator. The model is pre-trained on general headline generation, as~\cite{song2023general, yang2023fact},  for $2$ epochs with a peak learning rate~(LR) of $1e-4$ and cosine decay. Early stopping is applied within $5$ epochs for subsequent steps, with peak LRs set to $1e-3$, $1e-6$, and $1e-5$, respectively. Other details are consistent with prior work~\cite{yang2023fact}. The collaborative LLMs are Qwen2.5-72B-Instruct and GTE-Qwen2-7B-Instruct. 
We employ the Huggingface ROUGE pipeline
as prior work~\cite{song2023general} and re-report results for FPG~\cite{yang2023fact}. 
In the pairwise comparison evaluation of personalization performance, we use Qwen2.5-72B-Instruct as the judge to obtain the comprehensive assessment.
We use NVIDIA A800 80GB GPU for our experiments.

\begin{table}[tb]
\setlength{\abovecaptionskip}{0.75em} 
\setlength{\belowcaptionskip}{-0.5em}
\small
\centering
\caption{Performance of the Compared Baseline Methods.}
\label{tab:performance}
    \centering
    \resizebox{\linewidth}{!}{ 
    \begin{tabular}{clcccc}
        \toprule[1pt]
        \textbf{Types} & \textbf{Methods} & \textbf{ROUGE-1} & \textbf{ROUGE-2} & \textbf{ROUGE-L} & \textbf{Fact Scores}\\
        \midrule
        \multirow{4}{*}{Open-source LLMs} & FlanT5-large & 24.91 & 8.48 & 21.13 & 80.83 \\
         & FlanT5-XL & 27.73 & 10.08 & 23.23 & 84.16 \\
         & Qwen2.5-1.5B & 27.45 & 9.45 & 22.24 & 83.34 \\
         & Qwen2.5-7B & \underline{29.54} & \underline{10.47} & \underline{24.04} & \underline{88.67} \\[1pt]
        \midrule
        \multirow{4}{*}{LLMs API services} & GLM-4-Flash & 26.91 & 8.40 & 21.47 & 85.88 \\
         & GLM-4-Air & 27.45 & 8.89 & 22.03 & 88.02 \\
         & DeepSeek-V2.5 & 28.27 & 9.55 & 22.60 & 86.90 \\
         & GPT-4o & \underline{29.58} & \underline{11.03} & \underline{24.26} & \underline{90.11} \\[1pt]
        \midrule
        \multirow{7}{*}{Personalized models} & PENS-NRMS & 26.15 & 9.37 & 21.03 & 50.73 \\
         & PENS-NAML & 28.01 & 10.72 & 22.24 & 50.16 \\
         & PNG & 28.78 & 11.27 & 22.39 & 51.23 \\
         & EUI-PENS & 32.34 & 13.93 & 26.90 & \text{NA} \\
         & FPG & 33.06 & 13.76 & 26.78 & \underline{89.55} \\
         & GTP & \underline{33.84} & \underline{14.23} & \underline{27.85} & \text{NA} \\[1pt]
         \cdashline{2-6}
         \addlinespace[2pt]
         & $\textbf{SCAPE}_{\footnotesize \text{ours}}$ & $\mathbf{34.26}^{*}$ & $\mathbf{14.79}^{*}$ & $\mathbf{28.36}^{*}$ & $\mathbf{92.36}^{*}$ \\
        \bottomrule[1pt]
        \end{tabular}
    }
\\[2pt]
\noindent \footnotesize 
\textit{The symbol * denotes the significance level with $p \leq 0.05$. \textbf{Bold} font indicates the best-performing method. \underline{Underline} indicates the second-best results in the group.}
\vspace{-0.5em}
\end{table}

\begin{figure}[tb]
\setlength{\abovecaptionskip}{0.5em} 
  \centering
  \includegraphics[width=\linewidth]{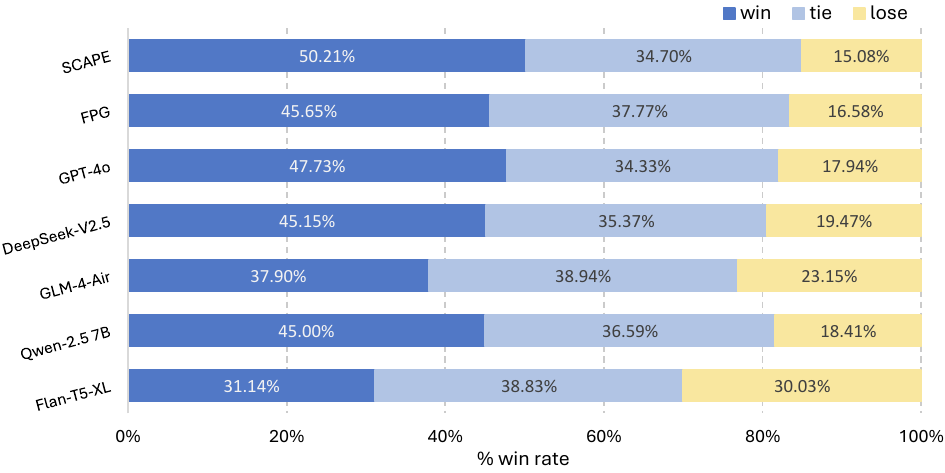}
  \caption{Win Rates in Personalization Evaluation.}
  \Description{The illustration of personalization evaluation Win Rates of customized LLMs for headline generation.}
  \label{fig:personalization}
  \vspace{-0.5em}
\end{figure}

\subsection{Results and Analysis}
Table~\ref{tab:performance} compares SCAPE with baseline methods, demonstrating its superior performance and setting a new benchmark for personalized headline generation in SOTA results. SCAPE significantly outperforms other methods in informativeness and fluency. This result highlights that SCAPE better improves content coverage through its stylistic-content-aware user modeling via LLM collaboration. Furthermore, SCAPE emphasizes the importance of balancing long- and short-term content interests and stylistic preferences to improve the factual consistency of personalized headlines.

Figure~\ref{fig:personalization} shows the win rates of personalization evaluation using LLMs with customized prompts as judges. 
While GPT-4o shows higher win rates than the personalized model FPG, LLMs still struggle to generate personalized headlines based on prompt engineering. This underscores the challenge posed by the complexity and diversity of user's clicks preferences. 
Our SCAPE framework takes a crucial step forward in personalized headlines generation by considering both user-oriented content interests and modeling linguistic style preferences in users' historical clicks.
\section{Conclusion}
In this paper, we propose SCAPE, a novel framework to tackle the challenges of insufficiently user preference modeling in personalized headline generation.
SCAPE introduces a headline inference module that extracts the inherent stylistic features and content interests from news headlines without explicit supervision.
A hierarchical gated fusion mechanism is further introduced to dynamically combine both long- and short-term interests for panoramic user modeling, which then guides the headline generator to produce stylistic-content aware personalized headlines.
Extensive experiments on the PENS dataset show that SCAPE sets a new state-of-the-art benchmark for personalized headline generation.


\begin{acks}
\small
The research work supported by National Key R\&D Plan No.2022YFC3303303, the National Natural Science Foundation of China under Grant (No. U2436209, 62476263).
Xiang Ao is also supported by the Project of Youth Innovation Promotion Association CAS, Beijing Nova Program 20230484430, the Innovation Funding of ICT, CAS under Grant No. E461060.
\end{acks}

\bibliographystyle{ACM-Reference-Format}
\balance
\bibliography{sample-base-full}


\begin{thebibliography}{22}


\ifx \showCODEN    \undefined \def \showCODEN     #1{\unskip}     \fi
\ifx \showISBNx    \undefined \def \showISBNx     #1{\unskip}     \fi
\ifx \showISBNxiii \undefined \def \showISBNxiii  #1{\unskip}     \fi
\ifx \showISSN     \undefined \def \showISSN      #1{\unskip}     \fi
\ifx \showLCCN     \undefined \def \showLCCN      #1{\unskip}     \fi
\ifx \shownote     \undefined \def \shownote      #1{#1}          \fi
\ifx \showarticletitle \undefined \def \showarticletitle #1{#1}   \fi
\ifx \showURL      \undefined \def \showURL       {\relax}        \fi
\providecommand\bibfield[2]{#2}
\providecommand\bibinfo[2]{#2}
\providecommand\natexlab[1]{#1}
\providecommand\showeprint[2][]{arXiv:#2}

\bibitem[Ao et~al\mbox{.}(2023)]%
        {ao2023put}
\bibfield{author}{\bibinfo{person}{Xiang Ao}, \bibinfo{person}{Ling Luo}, \bibinfo{person}{Xiting Wang}, {et~al\mbox{.}}} \bibinfo{year}{2023}\natexlab{}.
\newblock \showarticletitle{Put Your Voice on Stage: Personalized Headline Generation for News Articles}.
\newblock \bibinfo{journal}{\emph{ACM TKDD}} (\bibinfo{year}{2023}).
\newblock


\bibitem[Ao et~al\mbox{.}(2021)]%
        {ao2021pens}
\bibfield{author}{\bibinfo{person}{Xiang Ao}, \bibinfo{person}{Xiting Wang}, \bibinfo{person}{Ling Luo}, \bibinfo{person}{Ying Qiao}, \bibinfo{person}{Qing He}, {and} \bibinfo{person}{Xing Xie}.} \bibinfo{year}{2021}\natexlab{}.
\newblock \showarticletitle{PENS: A Dataset and Generic Framework for Personalized News Headline Generation}. In \bibinfo{booktitle}{\emph{Proc. of ACL 2021}}.
\newblock


\bibitem[Bell(1991)]%
        {bell1991language}
\bibfield{author}{\bibinfo{person}{Allan Bell}.} \bibinfo{year}{1991}\natexlab{}.
\newblock \bibinfo{booktitle}{\emph{The language of news media}}.
\newblock \bibinfo{publisher}{Blackwell Oxford}.
\newblock


\bibitem[Chung et~al\mbox{.}(2024)]%
        {chung2024scaling}
\bibfield{author}{\bibinfo{person}{Hyung~Won Chung}, \bibinfo{person}{Le Hou}, \bibinfo{person}{Shayne Longpre}, \bibinfo{person}{Barret Zoph}, \bibinfo{person}{Yi Tay}, \bibinfo{person}{William Fedus}, \bibinfo{person}{Yunxuan Li}, \bibinfo{person}{Xuezhi Wang}, \bibinfo{person}{Mostafa Dehghani}, \bibinfo{person}{Siddhartha Brahma}, {et~al\mbox{.}}} \bibinfo{year}{2024}\natexlab{}.
\newblock \showarticletitle{Scaling instruction-finetuned language models}.
\newblock \bibinfo{journal}{\emph{JMLR}} (\bibinfo{year}{2024}).
\newblock


\bibitem[Fiske(2010)]%
        {fiske2010introduction}
\bibfield{author}{\bibinfo{person}{John Fiske}.} \bibinfo{year}{2010}\natexlab{}.
\newblock \bibinfo{booktitle}{\emph{Introduction to communication studies}}.
\newblock \bibinfo{publisher}{Routledge}.
\newblock


\bibitem[GLM et~al\mbox{.}(2024)]%
        {glm2024chatglm}
\bibfield{author}{\bibinfo{person}{Team GLM}, \bibinfo{person}{Aohan Zeng}, \bibinfo{person}{Bin Xu}, \bibinfo{person}{Bowen Wang}, \bibinfo{person}{Chenhui Zhang}, \bibinfo{person}{Da Yin}, \bibinfo{person}{Dan Zhang}, \bibinfo{person}{Diego Rojas}, \bibinfo{person}{Guanyu Feng}, \bibinfo{person}{Hanlin Zhao}, {et~al\mbox{.}}} \bibinfo{year}{2024}\natexlab{}.
\newblock \showarticletitle{Chatglm: A family of large language models from glm-130b to glm-4 all tools}.
\newblock \bibinfo{journal}{\emph{arXiv preprint arXiv:2406.12793}} (\bibinfo{year}{2024}).
\newblock


\bibitem[Gu et~al\mbox{.}(2020)]%
        {gu2020generating}
\bibfield{author}{\bibinfo{person}{Xiaotao Gu}, \bibinfo{person}{Yuning Mao}, \bibinfo{person}{Jiawei Han}, {et~al\mbox{.}}} \bibinfo{year}{2020}\natexlab{}.
\newblock \showarticletitle{Generating Representative Headlines for News Stories}. In \bibinfo{booktitle}{\emph{Proc. of The Web Conference 2020}}.
\newblock


\bibitem[Hurst et~al\mbox{.}(2024)]%
        {hurst2024gpt}
\bibfield{author}{\bibinfo{person}{Aaron Hurst}, \bibinfo{person}{Adam Lerer}, \bibinfo{person}{Adam~P Goucher}, \bibinfo{person}{Adam Perelman}, \bibinfo{person}{Aditya Ramesh}, \bibinfo{person}{Aidan Clark}, \bibinfo{person}{AJ Ostrow}, \bibinfo{person}{Akila Welihinda}, \bibinfo{person}{Alan Hayes}, \bibinfo{person}{Alec Radford}, {et~al\mbox{.}}} \bibinfo{year}{2024}\natexlab{}.
\newblock \showarticletitle{Gpt-4o system card}.
\newblock \bibinfo{journal}{\emph{arXiv preprint arXiv:2410.21276}} (\bibinfo{year}{2024}).
\newblock


\bibitem[Jin et~al\mbox{.}(2020)]%
        {jin2020hooks}
\bibfield{author}{\bibinfo{person}{Di Jin}, \bibinfo{person}{Zhijing Jin}, \bibinfo{person}{Joey~Tianyi Zhou}, {et~al\mbox{.}}} \bibinfo{year}{2020}\natexlab{}.
\newblock \showarticletitle{Hooks in the Headline: Learning to Generate Headlines with Controlled Styles}. In \bibinfo{booktitle}{\emph{Proc. of ACL 2020}}.
\newblock


\bibitem[Li et~al\mbox{.}(2021)]%
        {li2021style}
\bibfield{author}{\bibinfo{person}{Mingzhe Li}, \bibinfo{person}{Xiuying Chen}, \bibinfo{person}{Min Yang}, {et~al\mbox{.}}} \bibinfo{year}{2021}\natexlab{}.
\newblock \showarticletitle{Learning to Write Eye-Catching Headlines via Disentanglement}. In \bibinfo{booktitle}{\emph{Proc. of AAAI 2021}}.
\newblock


\bibitem[Liu et~al\mbox{.}(2024)]%
        {liu2024deepseek}
\bibfield{author}{\bibinfo{person}{Aixin Liu}, \bibinfo{person}{Bei Feng}, \bibinfo{person}{Bin Wang}, \bibinfo{person}{Bingxuan Wang}, \bibinfo{person}{Bo Liu}, \bibinfo{person}{Chenggang Zhao}, \bibinfo{person}{Chengqi Dengr}, \bibinfo{person}{Chong Ruan}, \bibinfo{person}{Damai Dai}, \bibinfo{person}{Daya Guo}, {et~al\mbox{.}}} \bibinfo{year}{2024}\natexlab{}.
\newblock \showarticletitle{Deepseek-v2: A strong, economical, and efficient mixture-of-experts language model}.
\newblock \bibinfo{journal}{\emph{arXiv preprint arXiv:2405.04434}} (\bibinfo{year}{2024}).
\newblock


\bibitem[Luo et~al\mbox{.}(2019)]%
        {luo2019reading}
\bibfield{author}{\bibinfo{person}{Ling Luo}, \bibinfo{person}{Xiang Ao}, \bibinfo{person}{Yan Song}, \bibinfo{person}{Feiyang Pan}, \bibinfo{person}{Min Yang}, {and} \bibinfo{person}{Qing He}.} \bibinfo{year}{2019}\natexlab{}.
\newblock \showarticletitle{Reading like HER: Human Reading Inspired Extractive Summarization}. In \bibinfo{booktitle}{\emph{Proc. of EMNLP-IJCNLP 2019}}.
\newblock


\bibitem[Peng et~al\mbox{.}(2024)]%
        {peng2024answer}
\bibfield{author}{\bibinfo{person}{Letian Peng}, \bibinfo{person}{Yuwei Zhang}, \bibinfo{person}{Zilong Wang}, \bibinfo{person}{Jayanth Srinivasa}, \bibinfo{person}{Gaowen Liu}, \bibinfo{person}{Zihan Wang}, {and} \bibinfo{person}{Jingbo Shang}.} \bibinfo{year}{2024}\natexlab{}.
\newblock \showarticletitle{Answer is All You Need: Instruction-following Text Embedding via Answering the Question}. In \bibinfo{booktitle}{\emph{Proc. of ACL 2024}}.
\newblock


\bibitem[Shu et~al\mbox{.}(2018)]%
        {shu2018deep}
\bibfield{author}{\bibinfo{person}{Kai Shu}, \bibinfo{person}{Suhang Wang}, \bibinfo{person}{Thai Le}, {et~al\mbox{.}}} \bibinfo{year}{2018}\natexlab{}.
\newblock \showarticletitle{Deep Headline Generation for Clickbait Detection}. In \bibinfo{booktitle}{\emph{Proc. of ICDM 2018}}.
\newblock


\bibitem[Song et~al\mbox{.}(2023)]%
        {song2023general}
\bibfield{author}{\bibinfo{person}{Yun-Zhu Song}, \bibinfo{person}{Yi-Syuan Chen}, \bibinfo{person}{Lu Wang}, {and} \bibinfo{person}{Hong-Han Shuai}.} \bibinfo{year}{2023}\natexlab{}.
\newblock \showarticletitle{General then Personal: Decoupling and Pre-training for Personalized Headline Generation}.
\newblock \bibinfo{journal}{\emph{TACL}} (\bibinfo{year}{2023}).
\newblock


\bibitem[Tan et~al\mbox{.}(2024)]%
        {tan2024enhancing}
\bibfield{author}{\bibinfo{person}{Xiaoyu Tan}, \bibinfo{person}{Leijun Cheng}, \bibinfo{person}{Xihe Qiu}, {et~al\mbox{.}}} \bibinfo{year}{2024}\natexlab{}.
\newblock \showarticletitle{Enhancing Personalized Headline Generation via Offline Goal-Conditioned Reinforcement Learning with Large Language Models}. In \bibinfo{booktitle}{\emph{Proc. of KDD 2024}}.
\newblock


\bibitem[Wang et~al\mbox{.}(2024)]%
        {wang2024large}
\bibfield{author}{\bibinfo{person}{Peiyi Wang}, \bibinfo{person}{Lei Li}, \bibinfo{person}{Liang Chen}, \bibinfo{person}{Zefan Cai}, \bibinfo{person}{Dawei Zhu}, \bibinfo{person}{Binghuai Lin}, \bibinfo{person}{Yunbo Cao}, \bibinfo{person}{Qi Liu}, \bibinfo{person}{Tianyu Liu}, {and} \bibinfo{person}{Zhifang Sui}.} \bibinfo{year}{2024}\natexlab{}.
\newblock \showarticletitle{Large language models are not fair evaluators}. In \bibinfo{booktitle}{\emph{Proc. of ACL 2024}}.
\newblock


\bibitem[Yang et~al\mbox{.}(2024)]%
        {qwen2}
\bibfield{author}{\bibinfo{person}{An Yang}, \bibinfo{person}{Baosong Yang}, \bibinfo{person}{Binyuan Hui}, \bibinfo{person}{Bo Zheng}, \bibinfo{person}{Bowen Yu}, \bibinfo{person}{Chang Zhou}, \bibinfo{person}{Chengpeng Li}, \bibinfo{person}{Chengyuan Li}, \bibinfo{person}{Dayiheng Liu}, \bibinfo{person}{Fei Huang}, {et~al\mbox{.}}} \bibinfo{year}{2024}\natexlab{}.
\newblock \showarticletitle{Qwen2 technical report}.
\newblock \bibinfo{journal}{\emph{arXiv preprint arXiv:2407.10671}} (\bibinfo{year}{2024}).
\newblock


\bibitem[Yang et~al\mbox{.}(2023)]%
        {yang2023fact}
\bibfield{author}{\bibinfo{person}{Zhao Yang}, \bibinfo{person}{Junhong Lian}, {and} \bibinfo{person}{Xiang Ao}.} \bibinfo{year}{2023}\natexlab{}.
\newblock \showarticletitle{Fact-Preserved Personalized News Headline Generation}. In \bibinfo{booktitle}{\emph{Proc. of ICDM 2023}}.
\newblock


\bibitem[Zhang and Yang(2023)]%
        {zhang2023mediahg}
\bibfield{author}{\bibinfo{person}{Boning Zhang} {and} \bibinfo{person}{Yang Yang}.} \bibinfo{year}{2023}\natexlab{}.
\newblock \showarticletitle{MediaHG: Rethinking Eye-catchy Features in Social Media Headline Generation}. In \bibinfo{booktitle}{\emph{Proc. of EMNLP 2023}}.
\newblock


\bibitem[Zhang et~al\mbox{.}(2022)]%
        {zhang2022personalized}
\bibfield{author}{\bibinfo{person}{Kui Zhang}, \bibinfo{person}{Guangquan Lu}, \bibinfo{person}{Guixian Zhang}, \bibinfo{person}{Zhi Lei}, {and} \bibinfo{person}{Lijuan Wu}.} \bibinfo{year}{2022}\natexlab{}.
\newblock \showarticletitle{Personalized Headline Generation with Enhanced User Interest Perception}. In \bibinfo{booktitle}{\emph{Proc. of ICANN 2022}}.
\newblock


\bibitem[Zheng et~al\mbox{.}(2022)]%
        {zheng2022disentangling}
\bibfield{author}{\bibinfo{person}{Yu Zheng}, \bibinfo{person}{Chen Gao}, \bibinfo{person}{Jianxin Chang}, \bibinfo{person}{Yanan Niu}, \bibinfo{person}{Yang Song}, \bibinfo{person}{Depeng Jin}, {and} \bibinfo{person}{Yong Li}.} \bibinfo{year}{2022}\natexlab{}.
\newblock \showarticletitle{Disentangling long and short-term interests for recommendation}. In \bibinfo{booktitle}{\emph{Proc. of The Web Conference 2022}}.
\newblock


\end{thebibliography}

\end{document}